# A COMPREHENSIVE STUDY ON MACHINE LEARNING METHODS TO INCREASE THE PREDICTION ACCURACY OF CLASSIFIERS AND REDUCE THE NUMBER OF MEDICAL TESTS REQUIRED TO DIAGNOSE ALZHEIMER'S DISEASE

Md. Sharifur Rahman[1] and Professor Girijesh Prasad[2]

School of Computing, Engineering, and Intelligent Systems, Ulster University, Northern Ireland, UK

[1]rahman-m11@ulster.ac.uk
[2]g.prasad@ulster.ac.uk

## ABSTRACT

*Alzheimer's patients gradually lose their ability to think, behave, and interact with others. Medical history, laboratory tests, daily activities, and personality changes can all be used to diagnose the disorder. A series of time-consuming and expensive tests are used to diagnose the illness. The most effective way to identify Alzheimer's disease is using a Random-forest classifier in this study, along with various other Machine Learning techniques. The main goal of this study is to fine-tune the classifier to detect illness with fewer tests while maintaining a reasonable disease discovery accuracy. We successfully identified the condition in almost 94% of cases using four of the thirty frequently utilized indicators.*

## 1. INTRODUCTION

Alzheimer's disease causes brain atrophy and cell death. In Alzheimer's disease, a person's mental, behavioral, and social skills gradually deteriorate, impairing their capacity to operate independently. This disease affects around 5.8 million persons in the USA [5]. The incidence of Alzheimer's disease is estimated to be between 60% and 70% among the approximately 50 million individuals worldwide who have dementia. People living with Alzheimer's often forget prior events. When memory loss is severe, everyday chores become impossible. Alzheimer's disease results in symptoms other than dementia, such as self-harming behavior, increased susceptibility to infections, including pneumonia, poor balance, injuries from falls, and difficulty controlling bowel and bladder function. 1 in 3 elders dies from Alzheimer's disease. Alzheimer's disease and prostate cancer increase the risk of death in the elderly [6]. Between 2000 and 2019, the mortality rate from Alzheimer's disease quadrupled, whereas heart disease, the leading cause of death, decreased. Seniors with Alzheimer's are twice as likely to pass away before age 80 as seniors without the condition at age 70. Alzheimer's disease affects 4 to 8% of persons aged 65 and older; however, some survive for up to 20 years after being diagnosed. Alzheimer's is the 6th most significant cause of death [3] [4]. With more patients, Alzheimer's has gained relevance, leading to more excellent studies. Another research used MRI image data and blood-based protein biomarkers to improve sickness detection. We used no-imaging data.

The Random-forest [RF] [1], [2] classifier will be utilized to build an Alzheimer's Machine Learning [8] model. First, we'll look for discrepancies in the data and repair them using the miss-forest [MF] [9] algorithm. We'll also try using the PCA approach to see which way is the most accurate. We'll fine-tune all relevant parameters in both circumstances to increase our model's accuracy. Later, we'll use Boruta [7] and the Dalex library [7] to look at features for feature selection. Our primary goal is to limit the number of tests that can be skipped.

## 2. DATASET DESCRIPTION

The Australian Alzheimer's Disease Neuroimaging Initiative (ADNI) has primarily focused on the Australian Imaging Biomarkers and Lifestyle Study of Aging (AIBL). The dataset of our study came from ADNI. The AIBL study team obtained data for an independent test. The AIBL non-imaging dataset has 862 persons tested at the beginning and stored their data as baseline (BL). Afterward, 400, 184, and 100 people registered for M18, M36, and M54 tests. The data comprised demographics (2), medical history (10), ApoE genotype (1), psychological/functional assessments (4), blood tests (12), and clinical diagnoses (1). The study's participants varied in age from 55 to 96. Despite the inconsistencies, no null data was found. Before data processing, we handled the class imbalance problem in the dependent variable class.

## 3. METHODOLOGY

Only clinical and cognitive datasets were used in this experiment. Separate files held the data. Inconsistent data made validating it difficult. First, we inspected each file and compared it to the data outline. This project's GUI was R Studio. R studio 2021.09.2 Build 382 and R 4.1.2. All the datasets were imported into R Studio as one file. We added a new feature, "age" to the dataset, which calculates the patient's age by subtracting birthdate from the exam date. We replaced any incorrect data with NA. We used Missforest [366] to regenerate the wrong data fields. 100 ntrees with ten iterations were used as Missforest variables. Default mtry was used for feature-based categorization. Superfluous columns, including personal information, were removed. After pre-processing the data, we split it 70/30 for training and testing. After pre-processing the data, we split it 70/30 for training and testing. AIBL data classified as HC⇒Healthy Control, MCI⇒Mild Cognitive Impairment, and AD⇒Alzheimer's disease. This study compared HC versus Non-HC (combining MCI and AD). Later, we checked class equality in the training dataset's dependent variable field. One class had 862 instances, and another had 320. To balance classes, we generated more synthetic data with SMOTE [10]. We used training data for all analysis and model construction, which were later tested and evaluated with the test data.

## 4. MODEL PREPARATION

We constructed three unique models and fine-tuned them to attain the highest accuracy. First, we built a Random Forest classification model. This method of supervised learning may be utilized for both classification and regression. Our experiment is a kind of supervised classification. As is often understood, more trees (ntree) [1] make a more robust Random Forest. It also creates decision trees from data samples and votes on the best. By averaging or integrating the outcomes of several decision trees, it prevents overfitting (mtry) [2]. Random forests are incredibly adaptable. No data scaling is required for the random forest. It is accurate even when data is not scaled. The Random Forest technique retains excellent accuracy even when substantial amounts of data are missing. Initially, we used 500 trees (ntree) without mtry optimization; afterward, we experimented with increasing the number of mtry and determined that mtry=5 was the ideal combination for this data. The results are detailed below. Then we used Random Forest with ntree=500 and 50 permutations of the Dalex library to assess feature significance. We used the Boruta approach with ntree=500 to investigate feature selection. After

20 rounds, we observed that just two features (neurologic and hepatic medical history) negatively impacted our prediction model's accuracy, while the others were essential. We ran the tests twice, once with scaled data and later without. Original data performed better than scaled data, as shown in Table 1.

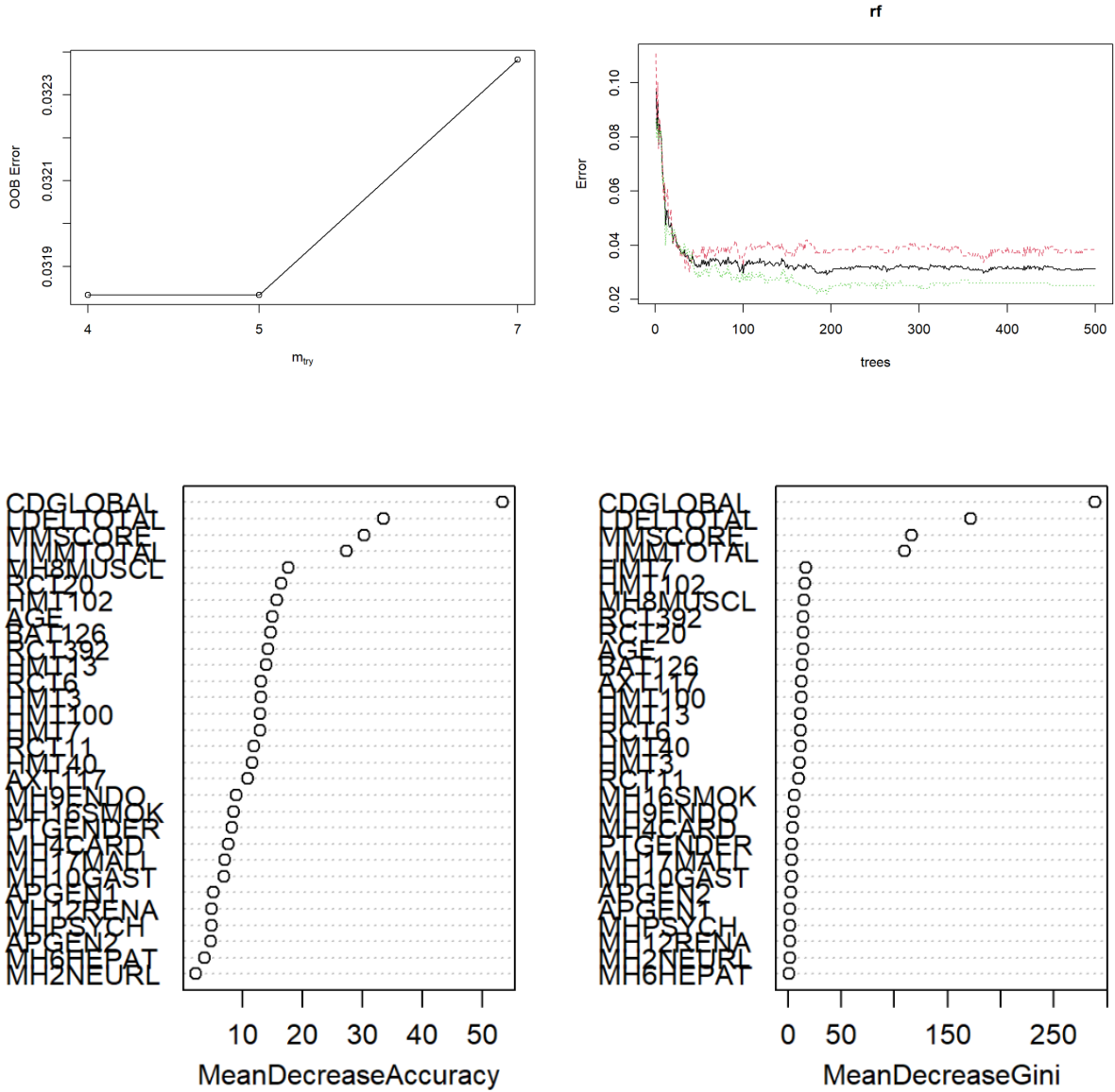

Figure 1. Tuning the Random Forest classifier for increased accuracy (increasing mtry) until OOB increases. The red curve in the rf plot denotes the Error for class 0, whereas the green curve denotes the Error for class 1. The black curve represents the OOB error.

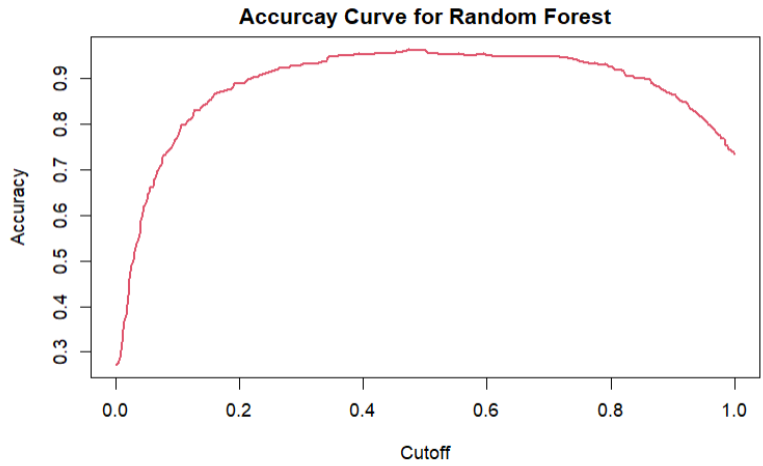

Figure 2. Model tweaking continued until the forecast accuracy rate increased. Tuning ceased when the test data accuracy rate dropped.

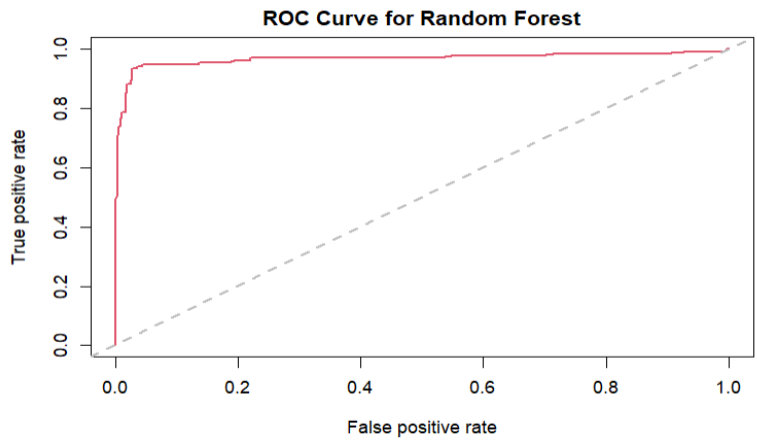

Figure 3. Receiver-Operating-Characteristic-(ROC)-curve-after-tuning.

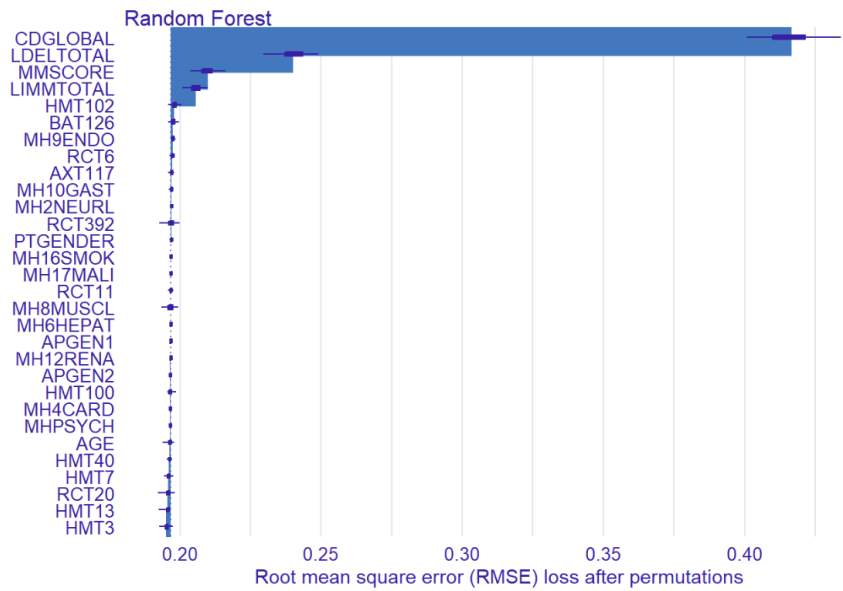

Figure 4. Using Dalex to measure features' importance.

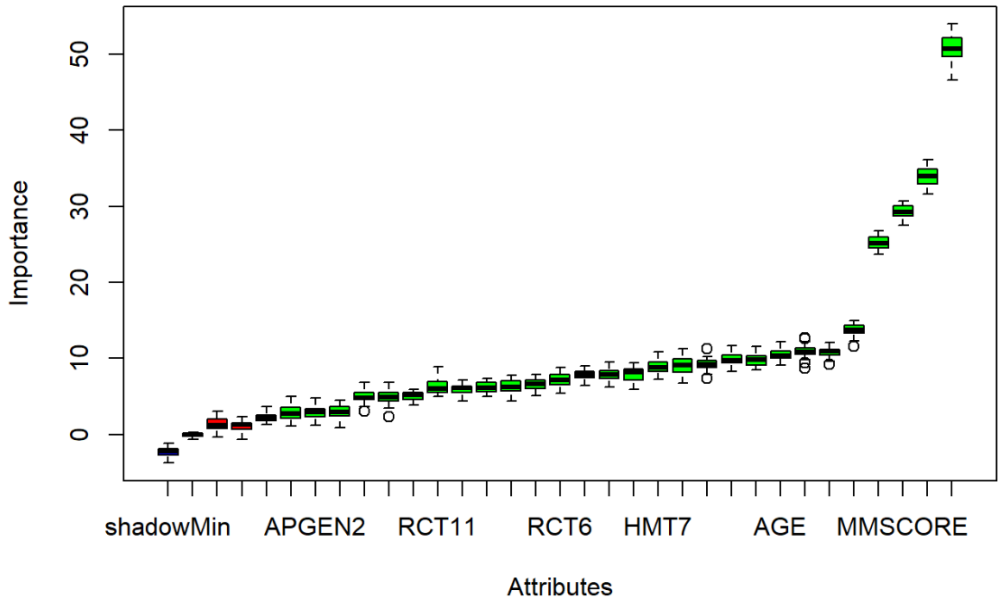

Figure 5. Confirming feature selection with the Boruta algorithm.

We built two classifier models using PCA. One used a predictor variable, and the other PCs. We tested the classifier's prediction accuracy by varying the number of PCs. We studied the trade-offs of adding PCs. In this experiment, PC=10 did better, with PC=28 being the best. In the last section, we created three classifier models of RF and tested their performance using medical history, neuropsychology evaluations, and blood analyses.

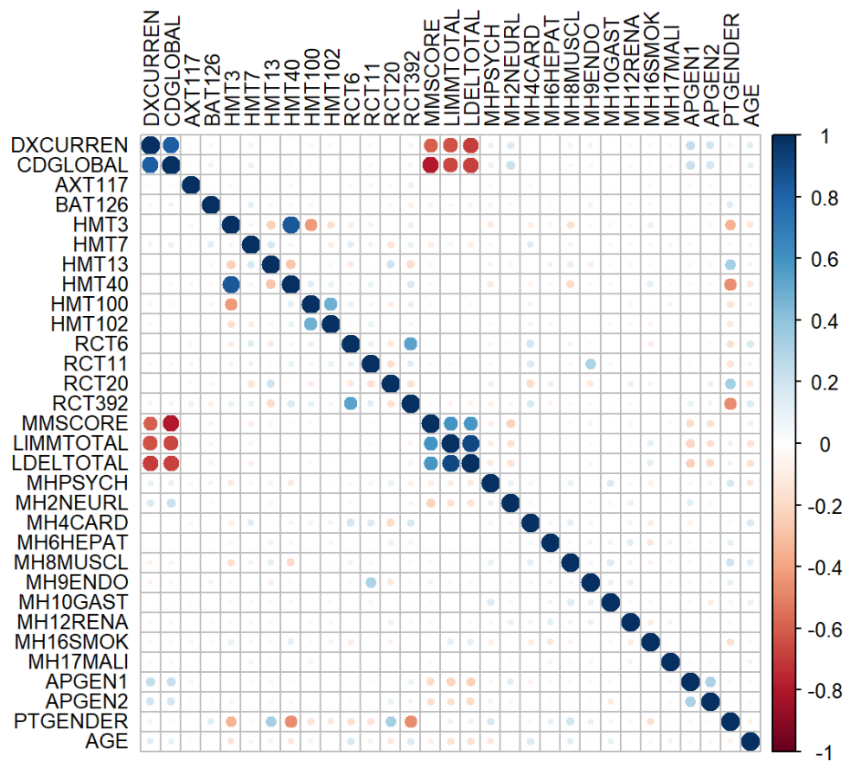

Figure 6. Correlation plot among predictor variables with PCA

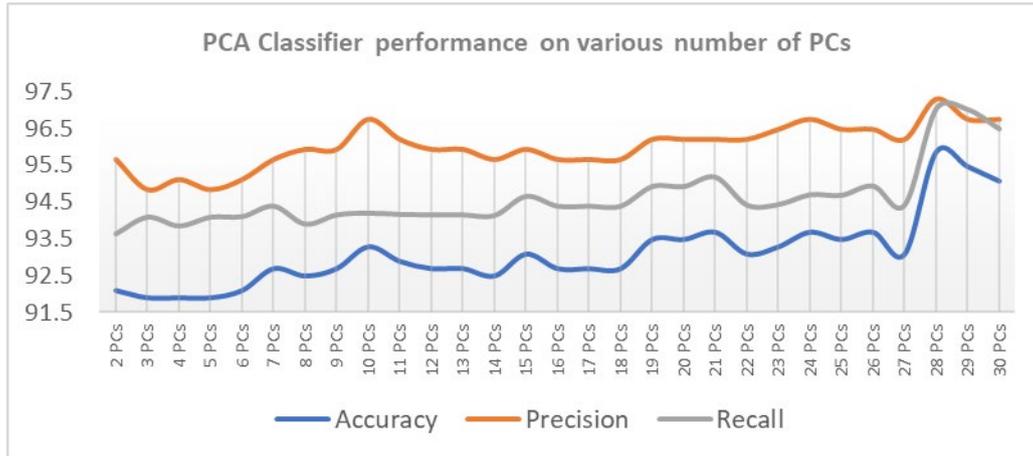

Figure 7. PCA shows 2 PCs are enough to get ~92% accuracy; to get ~93% accuracy, we need 10 PCs, and to get up to ~96% accuracy, we need 28 PCs which is a very high number. So, PCA was not worthy enough.

## 5. EVALUATION OF RESULTS

The diagnosis of Alzheimer's disease is our goal. Even if there is some uncertainty, it is always better to report the finding of an illness since it gives you a chance to take preventative action. Although both false-positive and false-negative outcomes are unexpected, false-positive is preferred over false-negative. We evaluated our classifier's performance using a confusion matrix. We computed accuracy, precision, and recall in this section. While accuracy is a more relevant metric for determining classifier performance, precision is equally critical. The performance score classifiers are shown in Table 1 from untuned to tuned.

Table 1. Classifier performance table

| Classifier | Data type | Accuracy | Precision | Recall |
|---|---|---|---|---|
| **Not tuned RF classifier** | Scaled data | 93.87 | 93.77 | 97.74 |
| **Tuned RF classifier** | Scaled data | 94.27 | 94.31 | 97.75 |
| **After feature selection with Dalex & Boruta** | Scaled data | 93.87 | 93.77 | 97.74 |
| **Not tuned RF classifier** | Original data | 95.65 | 97.02 | 97.02 |
| **Tuned RF classifier** | Original data | 96.05 | 97.29 | 97.29 |
| **After feature selection with Dalex & Boruta** | Original data | 95.45 | 97.02 | 96.76 |

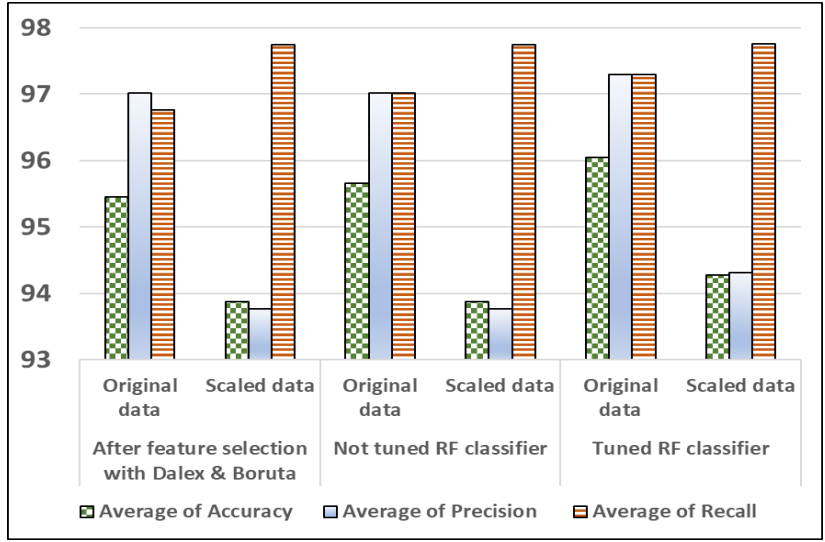

Figure 8. Performance graph for classifier models

We tested the random forest classifier with both scaled and unscaled data. Unscaled data was more accurate and precise than scaled data. Scaled data improves recall scores. Precision and accuracy are more important than recall. Table 2 shows performance score change using scaled data instead of unscaled data in classifier models.

Table 2. Assessment scores difference after scaling of classifier models

| Performance difference after scaling | | | |
|---|---|---|---|
| **Classifier** | **Accuracy** | **Precision** | **Recall** |
| **After feature selection with Dalex & Boruta** | -1.58 | -3.25 | 0.98 |
| **Tuned RF classifier** | -1.78 | -2.98 | 0.46 |
| **Not tuned RF classifier** | -1.78 | -3.25 | 0.72 |

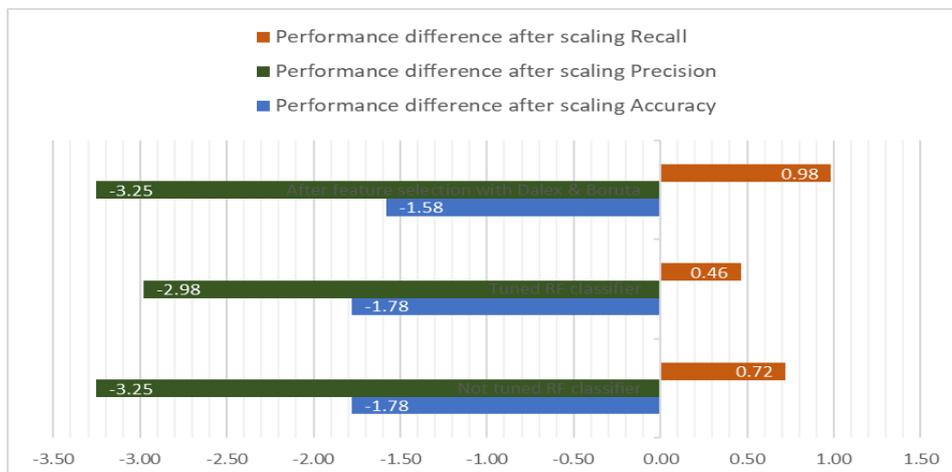

Figure 9. Classifier performance difference after scaling

We evaluated model performance on a particular data set. We discovered that we could utilize just Neuropsychology assessment data (clinical dementia rating, mini-mental state exam, logical memory immediate recall, and logical memory delayed recall test data) to get up to 93.68% accuracy and 95.66% precision. The performance score is shown in Table 3.

Table 3. Classifier performance over the specific datasets

| **Classifier based on** | **Accuracy** | **Precision** | **Recall** |
|---|---|---|---|
| **Medical history** | 67.79 | 71.54 | 81.99 |
| **Neuropsychology assessments** | 93.68 | 95.66 | 95.66 |
| **Blood analyses & ApoE genotypes** | 66.60 | 76.42 | 77.47 |

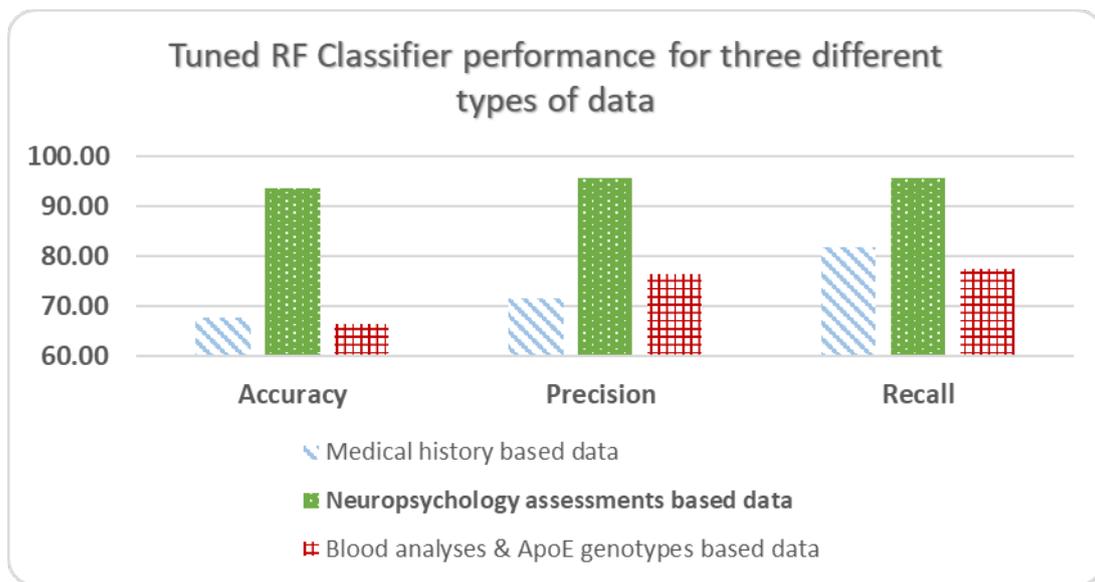

Figure 10. Neuropsychology assessment data set can get a satisfactory prediction rate.

## 6. CONCLUSIONS AND RECOMMENDATIONS FOR FUTURE WORK

Our data collection for the diagnosis of Alzheimer's disease was painstakingly designed and constructed by subject matter specialists. In addition, the results of our machine learning experiments indicated that practically all the features in the dataset are nearly weighted the same way in terms of how important they are. To reach the highest level of accuracy, we could only remove two test result columns from the prediction model. Even using PCA-based categorization, we are required to use more PCs than is typical for datasets of a similar size. After various tuning and testing, it was possible to conclude that we could obtain a respectable level of prediction accuracy and precision by concentrating on only neuropsychological tests, that is, four out of thirty tests. With 86% fewer data, accuracy only reduced by 2.37%, settling at 93.68%. Although the dataset comprises three classes, we restricted our experiment to two classes to accommodate all identified and potential disease cases in our experiment. In the future, there is space to research further, considering three classes.

# REFERENCES

[1] Breiman, L., 2001. Random forests. Machine learning, 45(1), pp.5-32.

[2] Random Forest algorithm (no date) www.javatpoint.com. Available at: https://www.javatpoint.com/machine-learning-random-forest-algorithm (Accessed: April 9, 2022).

[3] Xu JQ, Murphy SL, Kochanek KD, Arias E. Mortality in the United States, 2018. NCHS Data Brief; No. 355. Hyattsville, MD: National Center for Health Statistics. 2020.

[4] U.S. Department of Health and Human Services, Centers for Disease Control and Prevention, National Center for Health Statistics. CDC WONDER online database: About Underlying Cause of Death, 1999-2019. Available at: https://wonder.cdc. gov/ucd-icd10.html. Accessed December 28, 2021.

[5] Alzheimer's Disease and Dementia. Available at: https://www.alz.org/alzheimers-dementia/what-is-alzheimers (Accessed: April 27, 2022).

[6] Alzheimer's disease (no date) Mayo Clinic. Available at: https://www.mayoclinic.org/diseases-conditions/alzheimers-disease/symptoms-causes/syc-20350447 (Accessed: April 27, 2022).

[7] Kursa, M. B. and Rudnicki, W. R. (2010) "Feature Selection with the Boruta Package," Journal Of Statistical Software, 36(11).

[8] El Naqa, I. and Murphy, M. J. (2015) "What is machine learning?" in Machine Learning in Radiation Oncology. Cham: Springer International Publishing, pp. 3–11.

[9] Stekhoven, D. J. and Bühlmann, P. (2012) "MissForest--non-parametric missing value imputation for mixed-type data," Bioinformatics (Oxford, England), 28(1), pp. 112–118. doi: 10.1093/bioinformatics/btr597.

[10] Bhagat, R. C., and Patil, S. S. (2015) "Enhanced SMOTE algorithm for classification of imbalanced big-data using Random Forest," in 2015 IEEE International Advance Computing Conference (IACC). IEEE, pp. 403–408.

## Authors

**Md. Sharifur Rahman**

He is currently conducting data science research at Ulster University in the United Kingdom. Prior to this, he received master's and bachelor's degrees in computer science and engineering, as well as the Vice Chancellor's award. In addition, he holds an MBA. He is passionate about data mining and modelling in economic and medical research. He was driven to go deeper into data analysis after sixteen years of working in the telecom industry to integrate theoretical understanding with real-world applications to produce unique ideas that would help businesses and society.

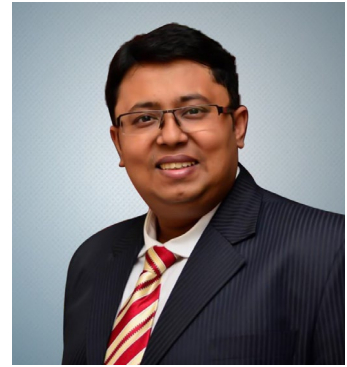

**Professor Girijesh Prasad**

He earned a BTech in Electrical Engineering from Regional Engineering College in Calicut, India, which is now the National Institute of Technology, and an MTech in Computer Science and Technology from the University of Roorkee, India, which is now the Indian Institute of Technology Roorkee, and a Ph.D. in Electrical Engineering from Queen's University of Belfast, UK, all in the same academic year. He is a Chartered Engineer, a Fellow of the IET, a Fellow of the Higher Education Academy, a Senior Member of the IEEE, and a founding member of the Technical Committee on Brain-Machine Interface Systems of the IEEE Systems, Man, and Cybernetics society. In 2017, he received the Senior Distinguished Research Fellowship from Ulster University and the International Academy of Physical Sciences (IAPS) India.

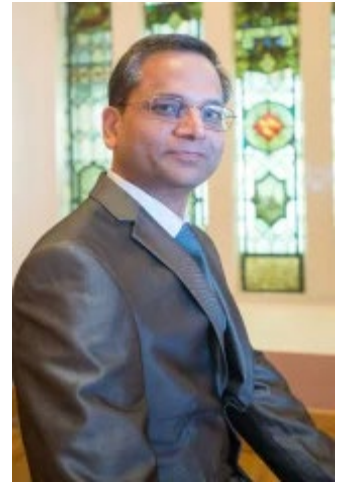

He studies assistive technologies, intelligent systems, data engineering, brain modeling, brain-computer interface (BCI), and neurorehabilitation. Under his direction, a neuro-rehab system with a robotic hand exoskeleton and an EEG/EEG-EMG based BCI has developed an advanced rehabilitation protocol with active physical and mental practice stages. This regimen has dramatically improved chronic stroke patients' quality of life in the UK and India. He published around 285 research papers in journals, edited volumes, and conference proceedings. He's supervised 22 PhDs. Invest Northern Ireland, the Department of Employment and Learning, the Royal Society, the Leverhulme Trust, UKIERI, UKRI, and Irish businesses have awarded him 18 research grants totaling over £10 million.